\documentclass[conference]{IEEEtran}
\IEEEoverridecommandlockouts

\usepackage{amsmath,amssymb,amsfonts}
\usepackage{graphicx}
\usepackage{textcomp}
\usepackage{xcolor}
\usepackage[utf8]{inputenc}
\usepackage{algorithm}
\usepackage{algpseudocode}

\usepackage[style=ieee]{biblatex}
\AtBeginBibliography{\small}
\begin{document}

\title{Flow-based path planning for multiple homogenous UAVs for outdoor formation-flying\\

\thanks{The authors of this work are employees of ANTS Aerial Systems Ltd, based in Dhaka, Bangladesh. This paper is an outcome of a drone light show project developed by the company.}
}

\makeatletter
\newcommand{\linebreakand}{%
  \end{@IEEEauthorhalign}
  \hfill\mbox{}\par
  \mbox{}\hfill\begin{@IEEEauthorhalign}
}
\makeatother

\author{
\IEEEauthorblockN{Mahmud Suhaimi Ibrahim}
\IEEEauthorblockA{\textit{System Engineer} \\
mutashi@gmail.com}

\and

\IEEEauthorblockN{Shantanu Rahman}
\IEEEauthorblockA{\textit{Software Engineer} \\
rahmanshantanu69@gmail.com}

\and

\IEEEauthorblockN{Muhammad Samin Hasan}
\IEEEauthorblockA{\textit{Chief Research Officer} \\
muhammadsaminhasan@gmail.com}

\linebreakand

\IEEEauthorblockN{Minhaj Uddin Ahmad}
\IEEEauthorblockA{\textit{System Engineer} \\
minhajuddin@iut-dhaka.edu}

\and

\IEEEauthorblockN{Abdullah Abrar}
\IEEEauthorblockA{\textit{Software Engineer} \\
abdullahabrar@iut-dhaka.edu}
}

\maketitle

\begin{abstract}
Collision-free path planning is the most crucial component in multi-UAV formation-flying (MFF). We use unlabeled homogenous quadcopters (UAVs) to demonstrate the use of a flow network to create complete (inter-UAV) collision-free paths. This procedure has three main parts: 1) Creating a flow network graph from physical GPS coordinates, 2) Finding a path of minimum cost (least distance) using any graph-based path-finding algorithm, and 3) Implementing the Ford-Fulkerson Method to find the paths with the maximum flow (no collision). Simulations of up to 64 UAVs were conducted for various formations, followed by a practical experiment with 3 quadcopters for testing physical plausibility and feasibility. The results of these tests show the efficacy of this method’s ability to produce safe, collision-free paths. 

\end{abstract}


\section{Introduction}
Multi-UAV path planning problems arise in various fields, including but not limited to, agriculture \cite{32-mogili2018review}, inspection \cite{33-hallermann2014visual, 34-metni2007uav}, entertainment \cite{35-kim2018survey}, and military applications \cite{37-wang2019reconnaissance}. The issue boils down to a group of UAVs trying to traverse from a set of initial locations to a set of goal locations without colliding with each other or environmental obstacles. For specific applications like outdoor multi-UAV formation flying (MFF), the UAVs need to form a certain shape. Often, in these cases, the UAVs do not need to be labeled. Meaning that the goal locations of the UAVs are independent from their identity.
 
A broad range of algorithms for multi-UAV path planning are available at the moment. We do not classify all of these approaches in this paper because our methodology as a whole does not fall into any pre-existing categories. An extensive summary of work and research on collision-free path-planning algorithms that exist to this day can be found in  \cite{5-huang2019collision} and \cite{8-madridano2021trajectory}. Having said that, we do provide a brief overview of a few methods and their limitations if they were to be used as standalone algorithms for collision-free path planning.
 
The classical path-finding methods such as A* in \cite{12-bentes2012dynamic}, RRT* in \cite{26-vcap2013multi} and their variations for single UAVs have been repurposed to work with multiple UAV systems. The main problem with these algorithms is that they do not have a collision avoidance system. The generated paths may overlap, and the speed of the UAVs would, in that case, need to be varied to avoid collisions. Nevertheless, their output could be used as an initiator for the Ford-Fulkerson Algorithm. 
 
Potential field algorithms emulate a field of energy where low-energy areas indicate lower chances of collision \cite{7-son2017lennard, 20-ma2016decentralized}. These approaches are useful in certain small, controlled spaces, but in a large and complex environment, they do not guarantee a complete solution. Improved variations of such techniques can be found in \cite{19-lopez2021path} with the Fast-Marching Square algorithm. These are still sub-optimal and will fail to find the solution in some instances.
 
Graph Neural Networks (GNN), another heavily researched method for multi-UAV path planning, is an extension of Convolutional Neural Networks (CNN) for graph data. To implement GNN, the path planning problem has to be formulated as a sequential decision-making problem. At every instant, each UAV takes an action with the objective of reaching its destination \cite{21-li2020graph}. It also requires the UAVs to constantly communicate with each other while traveling. This method allows the robots to have a certain level of autonomy and may not require mapping of the region to function. However, the drawback is that the computational complexity and cost of this method are high. Beyond the cost, there is no guarantee that it will reach the destination, which in cases of emergency response or light shows may be catastrophic \cite{4-madridano20203d}. 
 
Another more recent approach is to reinterpret the problem as a linear programming problem (LPP). This method is effective in a wide range of applications, but it is most commonly used in operations research or computer science. It might, however, just as readily be utilized for path planning. For example, \cite{24-yu2013multi} and \cite{25-yu2013planning} interpret the paths of multiple robots as flows in a flow network, which in essence is an LPP problem \cite{38-vanderbei2020linear}. Their solution, however, is limited to ground-based robots in a 2D environment.
 
Our approach extends the idea of using flow networks for path planning to aerial vehicles in an outdoor space; a 3D environment. It involves a process composed of several stages that utilize certain existing solutions and implement a few novel techniques. To show the robustness of our technique we have both simulated extreme collision-prone cases with numerous UAVs and conducted a practical experiment with quadcopters in an outdoor environment. 
 
The rest of the paper is divided into four sections. Section II describes the methodology and the theory behind our approach, followed by Section III, which gives detailed information about the components used for experimentation. Section IV is dedicated to the results and discussion of all the simulations and practical tests. Finally, Section V is a brief summary of the work and possible future improvements.

\section{Methodology}
Our approach to finding paths for a multi-UAV system involves two steps. First, we model the physical space as a flow network. A flow network is a directed graph where each edge has a capacity and receives a flow. Then we use an algorithm known as the Ford-Fulkerson algorithm to find the paths that provide maximum flow to the network.

\subsection{Directed Weighted Graphs}
A directed weighted graph, $G = (V, E)$, is composed of a set of \textit{vertices (nodes), V,} which are connected by a set of directed weighted \textit{edges, E}. For a pair of nodes $u, v \in V$, an edge $e = (u, v, w) \in E$, represents a connection from $u$ to $v$ with a weight, $w$. Concepts of \textit{flow} and the \textit{capacity} of an edge will be discussed in section \ref{flow-network-max}. 

\subsection{Bellman-Ford Algorithm}
The Bellman Ford Algorithm is a graph-based path-finding algorithm for finding the shortest path between a pair of nodes in a directed weighted graph. The starting node is called a source node, and the destination node is called the sink node. Bellman-Ford incrementally finds the shortest path between these two nodes. For any graph, $G = (V, E)$, let us define $dis[v]$, which is a mapping from $v$ to a number which is the cost of the shortest path available till now, and $from[v]$, which is also a mapping from $v$ to another vertex, which tells us from which $v$ was updated. We consider the outgoing edges of each vertex. It is trivial that, for any $edge(u, v, w)$, $dis[v] = min(dis[v], dis[u] + w)$, where the $dis[v]$ value is updated for any edge we set $from[v] = u$. This procedure is called \textit{incremental path relaxation}. The procedure is repeated $n-1$ times, where $n$ is the number of nodes. Backtracking the \textit{sink} node using flow mapping will give us the path of minimum cost from the \textit{source} to the \textit{sink}. Thus, a path is generated between the \textit{source} and the \textit{sink}.

\subsection{Flow Network and Maximum Flow Problem}
\label{flow-network-max}
A Flow network is a Directed Weighted Graph with some extra properties. The flow of an edge in a graph, $f(e)$, is a quantity that can represent various things depending on the situation. Our paper represents the number of UAVs that travel through an edge as flow. With that, we also have the capacity, $C(e)$, which is the maximum amount of flow allowed through an edge. Additionally, we define two special nodes: 
\begin{itemize}
    \item The \textit{source} node, $s$, which provides an infinite amount of flow and does not have any input edges. 
    \item The \textit{sink} node, $t$, which takes in an infinite amount of flow and does not have any outgoing edges. 
\end{itemize}
This concept will be crucial in solving our MFF problem, as we will see in section \ref{max-flow-prob-ford-fulk}.

\subsection{Flow Network generation}
To achieve our main objective of finding feasible paths for MFF, we first transform the physical space into a flow network. The process of this transformation will be discussed in detail in this section. To avoid inter-UAV collisions, we create a network where intersections between the paths are not possible. 

Initially, we take the start and goal positions of the UAVs and create a minimum bounding box that covers the smallest volume possible. We then divide the (3D) box into cubic cells with dimensions of \textit{d x d x d} meters. We choose the dimension based on the GPS or localization tolerance.

\begin{algorithm}
\begin{algorithmic}[1]
\caption{Finding Bounding Box}
\Function{FindBoundingBox}{$Positions$}
\State $x_{min} \gets \infty$
\State $y_{min} \gets \infty$
\State $z_{min} \gets \infty$
\State $x_{max} \gets 0$
\State $y_{max} \gets 0$
\State $z_{max} \gets 0$
\For{$point \gets points$}
    \State $x_{min} \gets \Call{Minimum}{x_{min}, point_x}$
    \State $x_{max} \gets \Call{Maximum}{x_{min}, point_x}$
    
    \State $y_{min} \gets \Call{Minimum}{y_{min}, point_y}$
    \State $y_{max} \gets \Call{Maximum}{y_{min}, point_y}$
    
    \State $z_{min} \gets \Call{Minimum}{z_{min}, point_z}$
    \State $z_{max} \gets \Call{Maximum}{z_{min}, point_z}$
\EndFor
\State \Return $\{ x_{max} - x_{min}, y_{max} - y_{min}, z_{max} - z_{min} \}$
\EndFunction
\end{algorithmic}
\end{algorithm}

\begin{algorithm}
\begin{algorithmic}[1]
\caption{Sub Dividing the Box into Grid}
\Function{SubDivideBoxIntoGrid}{$BoxSize$, $size$}
\State $X \gets BoxSize_x / size$
\State $Y \gets BoxSize_y / size$
\State $Z \gets BoxSize_z / size$
\State \Return [X, Y, Z]
\EndFunction
\end{algorithmic}
\end{algorithm}

Secondly, we further process the resulting grid to take into account static obstacles, free spaces, and valid spaces. After that, we are able to generate a Space Graph ($G_s$), shown in Fig. \ref{fig:space-graph}. $G_s$ is an undirected weighted graph where the nodes of the graph correspond to the cells in the grid and the edges between any two nodes, $n_a$ and $n_b$ are the paths available to the UAVs from $n_a$ to $n_b$ and vice-versa. The weight of the edge, in most cases, is the Manhattan distance between their corresponding grid cells. If, however, the movement is vertical, the cost is $H$ times the normal Manhattan distance. This is to discourage the UAVs from taking unnecessary vertical paths. If a cell in the grid contains a static obstacle, there will be no edges to that node and thus the UAVs will not be able to move into that node. Let us define nodes $n_{si}$ as being nodes that contain our UAVs' starting positions, and $n_{ti}$ as being the nodes that contain our goal positions.

\begin{algorithm}
\begin{algorithmic}[1]
\caption{Creating the Space Graph}
\Function{CreateSpaceGraph}{$GridSize$}
\State $N \gets \varnothing$
\State $E \gets \varnothing$
\State $H \gets 5$ \Comment{This is the vertical movement cost multiplier}
\State $directions \gets \varnothing$
\State $directions \gets directions \cup \{\{1, 0, 0\}\}$
\State $directions \gets directions \cup \{\{-1, 0, 0\}\}$
\State $directions \gets directions \cup \{\{0, 1, 0\}\}$
\State $directions \gets directions \cup \{\{0, -1, 0\}\}$
\State $directions \gets directions \cup \{\{0, 0, 1\}\}$
\State $directions \gets directions \cup \{\{0, 0, -1\}\}$
\For{$x \gets [0..GridSize[0]]$}
    \For{$y \gets [0..GridSize[1]]$}
        \For{$z \gets [0..GridSize[2]]$}
            \State $N = N \cup \Call{GetNode}{[x, y, z]}$
        \EndFor
    \EndFor
\EndFor
\State
\For{$x \gets [0..GridSize[0]]$}
    \For{$y \gets [0..GridSize[1]]$}
        \For{$z \gets [0..GridSize[2]]$}
            \For{$dir \gets directions$}
                \State $N_a \gets \Call{GetNode}{[x, y z]}$
                \State $N_b \gets \Call{GetNode}{[x, y z] + dir}$
                \State $mod \gets dir$
                \State $mod[2] \gets mod[2] \cdot H$
                \State $D \gets \Call{ManhattanDistance}{mod}$
                \State $E \gets E \cup \{\{ N_a, N_b, D \}\}$
            \EndFor
        \EndFor
    \EndFor
\EndFor
\State $G_{s} \gets (N, E)$
\State \Return $G_{s}$
\EndFunction
\State
\end{algorithmic}
\end{algorithm}

\begin{figure}[htbp]
\centering
    \includegraphics[width=\linewidth]{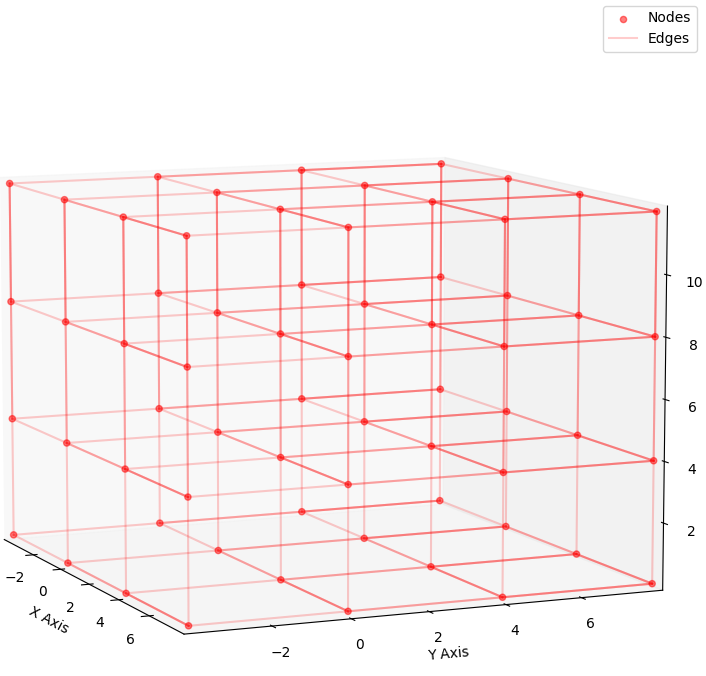}
    \caption{Space Graph showing the nodes and edges}
    \label{fig:space-graph}
\end{figure}

Thirdly, we generate another graph, the State Graph ($G_x$), as a directed weighted graph. For each node in $G_s$, let there be two types of nodes in $G_x$ called an entrance node and an exit node. We then define a directed edge from the entrance node to the exit node. For the edge between nodes $n_a$ and $n_b$ in $G_s$, we define two new directed edges from $n_{ax}$  to $n_{be}$ and from $n_{bx}$ to $n_{ae}$, shown in Fig. \ref{fig:pair-of-connected}, where $x$ means exit and $e$ means entrance. One of the paths will automatically become illegal, as will be shown later in the implementation of the Ford-Fulkerson algorithm.

\begin{algorithm}
\begin{algorithmic}[1]
\caption{Create State Graph}

\Function{CreateStateGraph}{$G_{s}, starts, goals$}
\State $N \gets \varnothing$
\State $E \gets \varnothing$
\State $C \gets \varnothing$

\For {$edge \gets G_{s}$}
    \State $n_a, n_b, d \gets edge$
    \State $n_{ae} \gets \Call{EntryNode}{n_a}$
    \State $n_{ax} \gets \Call{ExitNode}{n_a}$
    \State $n_{be} \gets \Call{EntryNode}{n_b}$
    \State $n_{bx} \gets \Call{ExitNode}{n_b}$
    \State $N \gets N \cup \{ n_{ae}, n_{ax}, n_{be}, n_{bx} \}$
    \State $NewEdges \gets \{ $
    \State $\;    \{ n_{ae}, n_{ax}, 0 \}, $
    \State $\;    \{ n_{be}, n_{bx}, 0 \}, $
    \State $\;    \{ n_{ax}, n_{be}, d \}, $
    \State $\;    \{ n_{bx}, n_{ae}, d \} $
    \State $ \}$
    \State $E \gets E \cup NewEdges$
\EndFor

\State $N \gets N \cup \{s, t\}$

\For {$node \gets starts$}
    \State $entry \gets \Call{EntryNode}{node}$
    \State $E \gets E \cup \{\{s, entry, 0\}\}$
\EndFor

\For {$node \gets goals$}
    \State $exit \gets \Call{ExitNode}{node}$
    \State $E \gets E \cup \{\{exit, t, 0\}\}$
\EndFor

\For{$edge \gets E$}
    \State $\Call{C}{edge} \gets 1$ \Comment{One UAV can cross one path}
\EndFor

\State $G_{x} \gets (N, E, C)$
\State \Return $G_{x}$

\EndFunction
\end{algorithmic}
\end{algorithm}

\begin{algorithm}
\begin{algorithmic}[1]
\caption{Whole Solution to MFF}
\Procedure{MFFSolver}{}
    \State $Start \gets \textnormal{the starting positions of the UAVs}$
    \State $Goal \gets \textnormal{the goal locations of the UAVs}$
    \State $MinBox \gets \Call{FindBoundingBox}{Start + Goal}$
    \State $Grid \gets \Call{SubDivideBoxIntoGrid}{MinBox}$
    \State $G_{s} \gets \Call{CreateSpaceGraph}{Grid}$
    \State $G_{x} \gets \Call{CreateStateGraph}{G_{s}}$
    \State $Paths \gets \Call{FordFulkerSon}{G_{x}}$
    \State $WayPoints \gets \Call{SamplePoints}{Paths, \#Waypoints}$
\EndProcedure
\end{algorithmic}
\end{algorithm}

\begin{figure}[htbp]
    \centering
    \includegraphics[width=\linewidth]{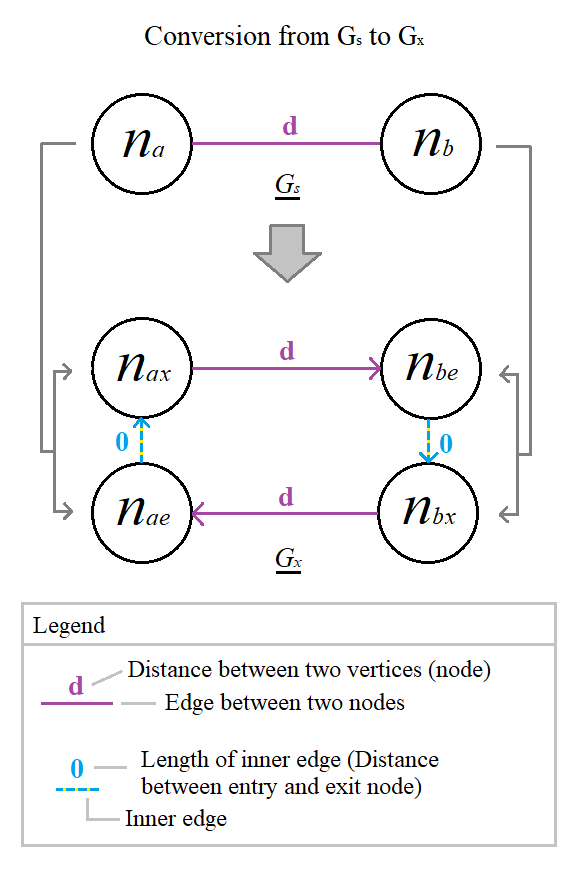}
    \caption{A pair of connected nodes in $G_s$ getting converted into Two pairs of nodes and interconnected edges in $G_x$}
    \label{fig:pair-of-connected}
\end{figure}

Finally, we set two abstract nodes, which do not correspond to any physical space, a \textit{virtual source node}, $s_v$, and a \textit{virtual sink node}, $t_v$, and we set a directed edge from the virtual source to each of the entrances of the starting nodes, $n_{sie}$ and a directed edge for each of the exits of the goal nodes, $n_{tix}$ to the \textit{virtual sink}. We set all the edges to have a capacity of 1, where the capacity of an edge is the number of times a UAV is able to move through that edge. We claim that the solution to the Maximum Flow Problem of this $G_x$ (flow network) will correspond to our collision-free paths in the grid space.

\subsection{Maximum Flow Problem and Ford Fulkerson Algorithm}
\label{max-flow-prob-ford-fulk}
The maximum flow problem asks a very simple question. Given a flow network, what is the maximum amount of flow that can be passed through the network from the $source$ node to the $sink$ node? In our case, we need to find how many UAVs we can send through the network so that no two UAVs occupy the same node at any instant.

We can interpret the path of a UAV as corresponding to a flow from the source node to the sink node. We guarantee that no UAV can occupy the same node concurrently by setting the capacity of the edges to 1. This is especially relevant on the inner edges between the entry and exit nodes, represented in Fig. \ref{fig:pair-of-connected} by the blue-yellow lines. As long as a UAV at any point moves to that node’s corresponding grid cell, the flow in the inner edge is set to be 1, which fulfills its capacity. By doing so, any possibility of collision is avoided. An explanation of this is given in Section \ref{section:conflict}.

In $G_x$, the maximum flow will always correspond to the number of total UAVs if and only if each UAV can pass through the network using a set of edges from the source to the sink in which the total flow is 1. In that case, there exists a set of collision-free paths for each UAV. The real paths of the UAVs will start from the node next to the sources and end just before the sinks, which in our case are conveniently the starting nodes and the ending nodes.

We use the Ford Fulkerson Algorithm \cite{30-ford1956maximal} to find a solution to the maximum flow problem in $G_x$. The algorithm depends upon three important ideas of residual networks, augmenting paths, and cuts \cite{31-cormen2022introduction}. The residual graph, $G_f$, is important for augmenting paths to find maximum flow. 

In a flow network, the maximum flow an edge can handle is the capacity of that edge, $C(e)$. The residual capacity of an edge, $R(e)$ is the maximum flow an edge can handle in the residual graph, $G_f$, therefore, for any edge, $R(e) = C(e) - f(e)$. It must also be assumed that the flow coming into a node must equal the flow out of a node, with the exception of the source and sink nodes. Thus, for any node $n \in V - \{s, t\}:$
\begin{equation}
    \sum_{e \in E: n \in e}{f(e)} = 0
\end{equation}

Therefore, it can be deduced that the flow mapping is skew-symmetric. So, for any two nodes $n_a, n_b \in V:$
\begin{equation}
    f(\{n_a, n_b\} = -f(\{n_b, n_a\})
\end{equation}

\begin{figure}[htbp]
    \centering
    \includegraphics[width=0.7\columnwidth]{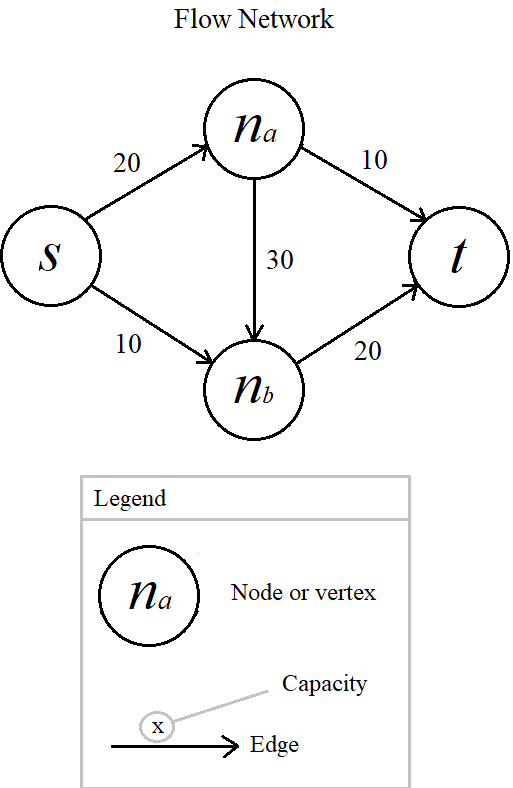}
    \caption{A flow network, G, with four nodes; s indicates the source, and t indicates the sink.}
    \label{fig:flownet}
\end{figure}

The procedure of the Ford-Fulkerson Algorithm is as follows:

\begin{enumerate}
    \item If a path from s - t is found, apply the bottleneck flow to it. The path in this case is $s \rightarrow n_a \rightarrow n_b \rightarrow t$. (Fig. \ref{fig:flownet})

    \item The bottleneck of the path, in this case, is 20, as this is the maximum flow possible in this path. Thus, you have to assign the flow to 20 for all edges in the path. (Fig. \ref{fig:directedpath})
    
\begin{figure}[htbp]
    \centering
    \includegraphics[width=0.7\columnwidth]{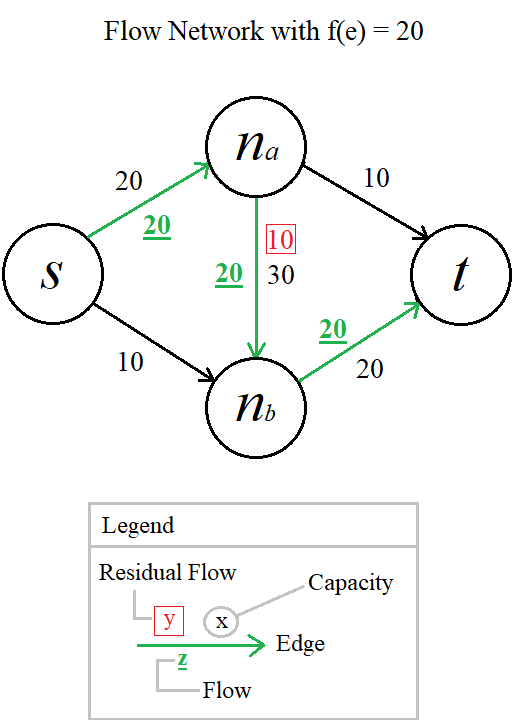}
    \caption{A directed path and the assigned minimum capacity on that path.}
    \label{fig:directedpath}
\end{figure}
    
    \item Build a residual graph $G_f$ as shown in Fig. \ref{fig:residual}. The edges with zero capacity are not shown as they are not feasible for traversal.

\begin{figure}[htbp]
    \centering
    \includegraphics[width=0.7\columnwidth]{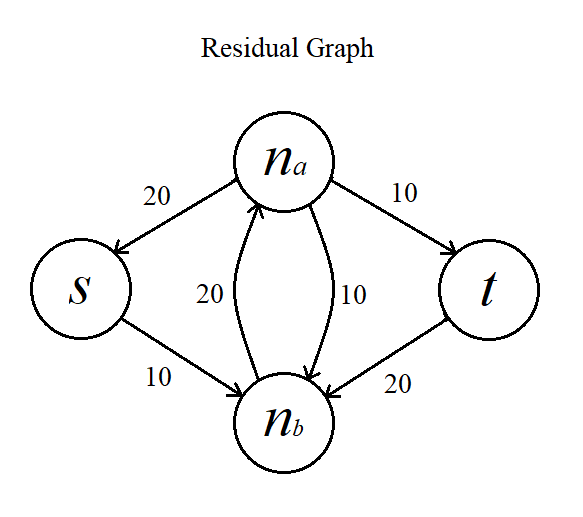}
    \caption{The residual graph $G_f$ with the new edges}
    \label{fig:residual}
\end{figure}

    \item For a path, $P$, in the residual graph, that goes from $s \rightarrow t$, let the bottleneck, $b$ be the minimum residual capacity on $P$. The path in this case is $s \rightarrow n_b \rightarrow n_a \rightarrow t$ where $b = 10$
    
    \item Augment the flow using the following equations, shown in Fig. \ref{fig:finalflow}
    \begin{itemize}
        \item If $e$ is a forward edge in $P$
        \begin{equation}
            f^{\prime}(e) = f(e) + b
        \end{equation}
        \item If $e$ is a backward edge in $P$
        \begin{equation}
            f^{\prime}(e) = f(e) - b
        \end{equation}
    \end{itemize}
    
\begin{figure}[htbp]
    \centering
    \includegraphics[width=0.7\columnwidth]{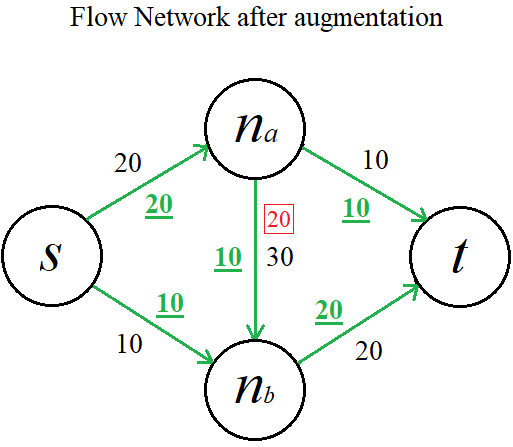}
    \caption{Final flow network with the augmented flow}
    \label{fig:finalflow}
\end{figure}

    \item Repeat steps 1-5 until the maximum flow is found.

\end{enumerate}

Note that there must be an $s \rightarrow t$ path in the residual graph for the algorithm to work. As shown in Fig. \ref{fig:nopath}, there is no path; hence the maximum flow is zero.

\begin{figure}[htbp]
    \centering
    \includegraphics[width=0.7\columnwidth]{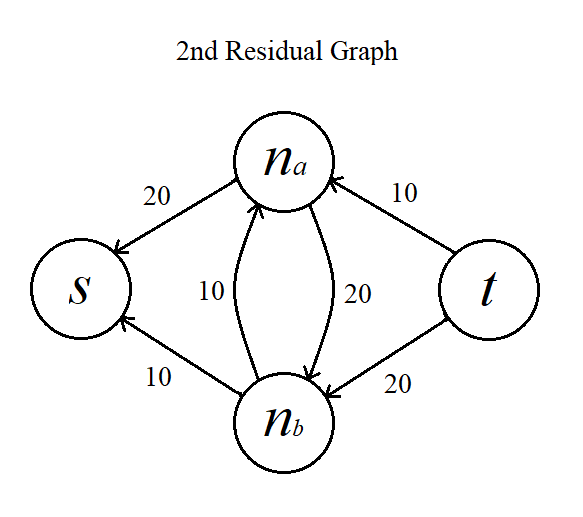}
    \caption{Final flow network with the augmented flow}
    \label{fig:nopath}
\end{figure}

\subsection{Conflict resolution using the Ford-Fulkerson Algorithm}
\label{section:conflict}
The Ford-Fulkerson algorithm simply augments the path so that the path found in Bellman-Ford is always \textit{feasible}, which in our case means collision-free. The algorithm also ensures the minimum cost is maintained. The complete proof of the algorithm is outside the scope of this paper.

We can, however, see an example of how the algorithm works in a small subset of the graph. In Fig. \ref{fig:conflict-resolution}, we can see if the green path is chosen first, then the red path cannot be chosen, as the two paths will definitely intersect. However, the flow through the red path seems to be zero at first. That is why we need to consider the internal edges between the entry and the exit of the nodes. The capacity of that internal edge has already been fulfilled as one UAV has chosen to pass through that edge. Thus, in actuality, the red path has a bottleneck of zero. Hence, no future UAVs can pass through it, and collision is avoided. A simple solution of such flow network is shown in Fig. \ref{fig:state-graph}.

\begin{figure}
    \centering
    \includegraphics[width=\linewidth]{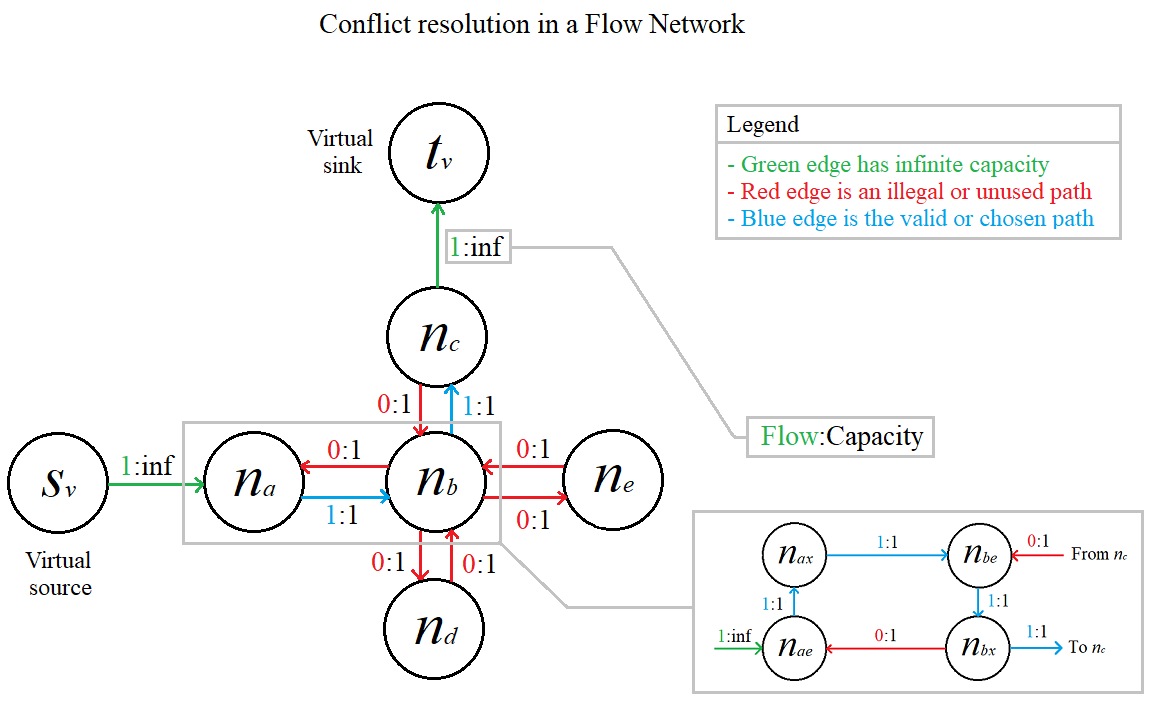}
    \caption{Conflict resolution in the Flow Network}
    \label{fig:conflict-resolution}
\end{figure}

\begin{figure}
    \centering
    \includegraphics[width=\linewidth]{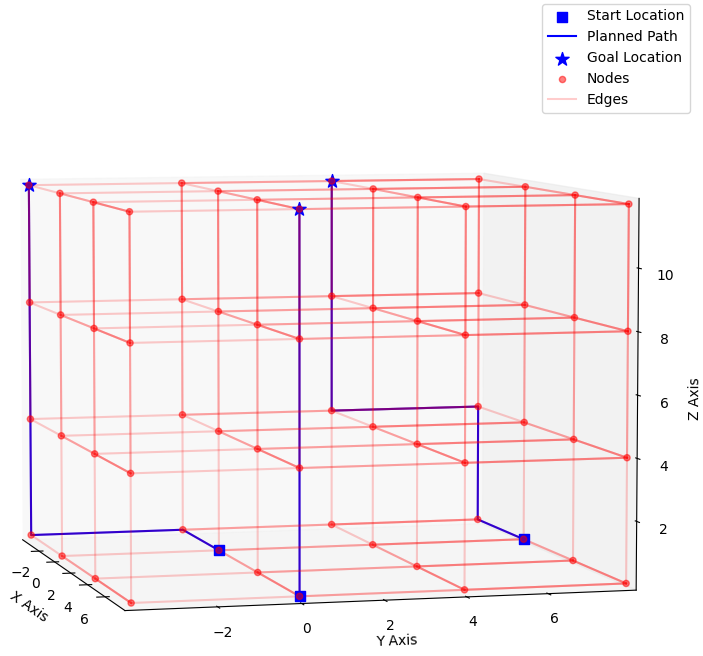}
    \caption{$G_x$ with a planned path for a certain formation}
    \label{fig:state-graph}
\end{figure}

\section{Experimental Setup}
\subsection{Simulations}
Before actual flights, we generated our planned paths and visualized them with different tools to ensure that our method for MFF actually works. The simulations were conducted on a computer with an 8th gen Intel Core i7 processor and 16GB of RAM. We used the Matplotlib library on Python 3.7 to visualize the patterns. The patterns were chosen in such a way that the collision-avoiding aspect of our algorithm is highlighted.

\subsection{Physical Experiment}
Making a quadcopter fly necessitates precise coordination of the motors. Dedicated flight controller hardware is typically used for this. The quadcopters in this work use a Pixhawk flight controller that has been flashed with Arducopter firmware. As the source codes and toolings of the Arducopter firmware are publicly available, it is a popular choice for hobbyists and research UAVs.

In this arrangement, all  the quadcopters are connected to a high-gain WiFi access point. The controlling computer and all of the quadcopters are connected to the same local area network. Most commercial WiFi routers come with a 255.255.255.0 (/24) subnet mask set. This means that at most 253 quadcopters and 1 controlling computer can be linked. But this can be configured to have a /23 or lower mask to allow more quadcopters on the network.

Arducopter's ``GUIDED" mode accepts commands from a remote controller or from a computer with a dedicated telemetry device. The flight controller can additionally take Mavlink \cite{2-koubaa2019micro} messages using its serial communication port. In this setup, Mavlink messages are sent to the flight controller using a wifi link and do not use any dedicated telemetry device. For MFF, the controlling computer sends necessary commands to the quadcopter's companion computer. The term ``companion computer” refers to a computer attached to the quadcopter unit itself, often a single-board computer (SBC) like the Raspberry Pi or Nvidia Jetson. The Raspberry Pi runs ``Mavproxy" software, which has the ability to forward Mavlink messages received from the controlling computer to the flight controller through a USB connection.

As for the mavlink messages, we use the dronekit library to send the Mavlink messages to the Raspberry Pi. These Mavlink messages are forwarded using the Mavproxy service on the Raspberry Pi \cite{2-koubaa2019micro} . Then we set the quadcopter to “GUIDED” mode and send the coordinates that they should go to. The Pixhawk manages the rest of the process of actually moving the quadcopter to that point.

The same computer that was used for the simulation of patterns was used to simulate the physical experiment as well. It was done in our own proprietary ground control application, shown in Fig. \ref{fig:gcs}.

\begin{figure}
    \centering
    \includegraphics[width=0.95\linewidth]{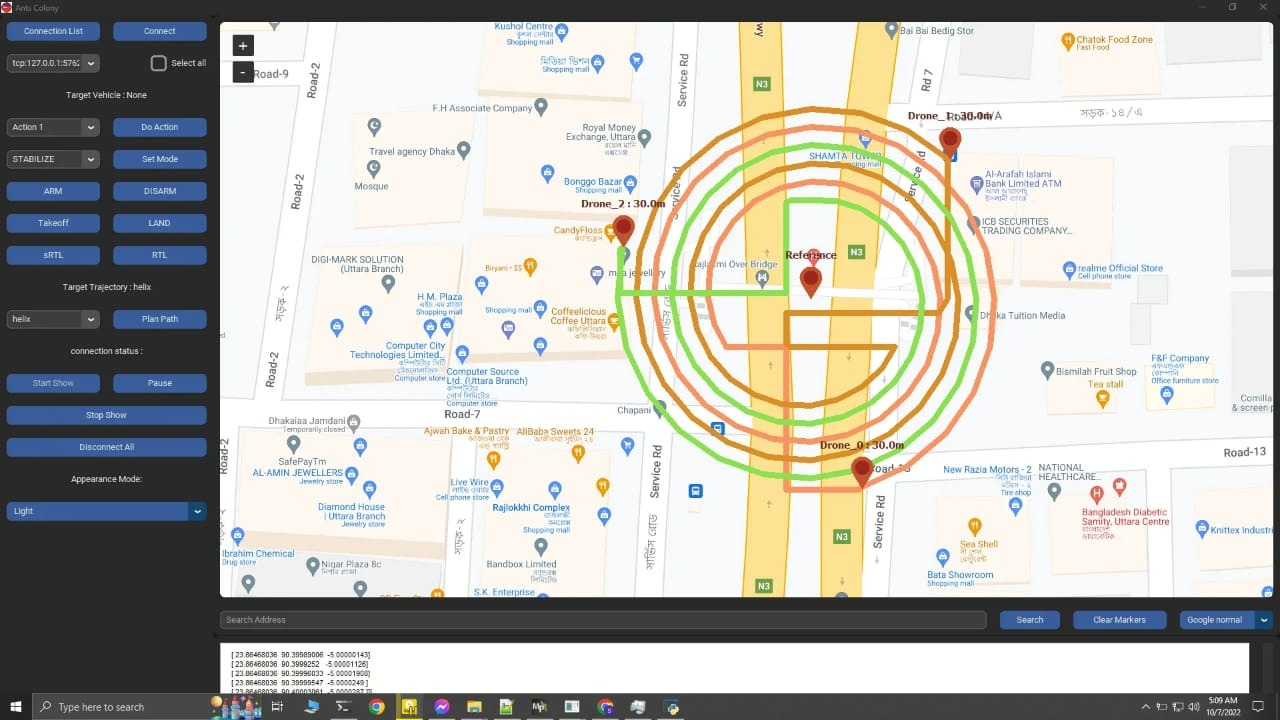}
    \caption{Our own ground control application running the simulation of the physical experiment.}
    \label{fig:gcs}
\end{figure}

\section{Results and Discussions}
\subsection{Simulations}
Figs. \ref{fig:ants}, \ref{fig:smiley}, and \ref{fig:cube} shows our simulation of several formations using UAVs. The red lines correspond to the paths taken by each of the UAVs. Although not clearly visible here in a two-dimensional plane, the paths do not intersect or overlap at any point and are void of collisions. As the complexity of the pattern increases, so do the number of UAVs needed for it. The number of nodes corresponds to the total volume of the grid. The computation time differs with different patterns. The 2D shape formations were computed within 2.5 minutes and the more complex, collision-prone 3D shapes took less than 5 minutes (See Table \ref{tbl:1}). The larger the number of connections, the longer the computation time. This is because our method has to first find a longer path and then resolve a greater number of conflicts to find collision-free paths. It has to be noted that the computation time has a larger correlation with the size of the grid than the number of UAVs.

\subsection{Physical test}
We implemented our algorithm and conducted successful real-life flight tests for MFF. In the experiment, quadcopters are first placed on a grid-like pattern on the XY plane. The starting location data is provided by the onboard GPS of the quadcopters. The median position of quadcopters is used as a local reference for coordinate transformation. The position data is converted from the Latitude, Longitude, and Altitude (LLA) geographic coordinate system to the local XYZ cartesian coordinate system.

After the conversion of the reference frame, a ``pattern'' is generated in the local frame. The initial location of all the quadcopters is used as a starting location, and points on the generated patterns are considered the goal locations.

Upon completion of path planning, a waypoint on each path was sent as a target location to the appropriate quadcopters in the formation. The above-mentioned step was followed until all the quadcopters had reached their destination.

The whole experiment was first conducted in our ground control application in order to check if our workflow is compatible with the Arducopter firmware. Upon completion of the simulation of the experiment, tests were carried out at the flying field, shown in Fig. \ref{fig:real1}. Three quadcopters completed their task to stay in formation and return without any collision. Fig. \ref{fig:real2} shows the quadcopters in mid-flight.

As a note, if a quadcopter lost connection or went rogue, it would come back using Smart Return To Launch (SRTL). In that case, the quadcopter just traverses the same path back that it used to go to its final known location. An unintended advantage of this policy is that even in the case of multiple quadcopters failing, the path followed back will always be collision-free.

\begin{table}
\caption{Simulations of multifarious formations and their respective computation times}
\label{tbl:1}
\begin{tabular}{ |c|c|c|c|c| } 
 \hline
 \textbf{Pattern} & \textbf{Number of} & \textbf{Nodes} & \textbf{Edges} & \textbf{Computation} \\
  & \textbf{UAVs} & & & \textbf{Time (s)} \\
 \hline
 ``ANTS" & 51 & 86940 & 254171 & 167.5 \\ 
 \hline
 Smiley Face & 44 & 101520 & 297982 & 148.5 \\
 \hline
 Cube & 64 & 146412 & 430500 & 299.4 \\
 \hline 
\end{tabular}
\end{table}

\begin{figure}[!ht]
    \centering
    \includegraphics[width=0.7\linewidth]{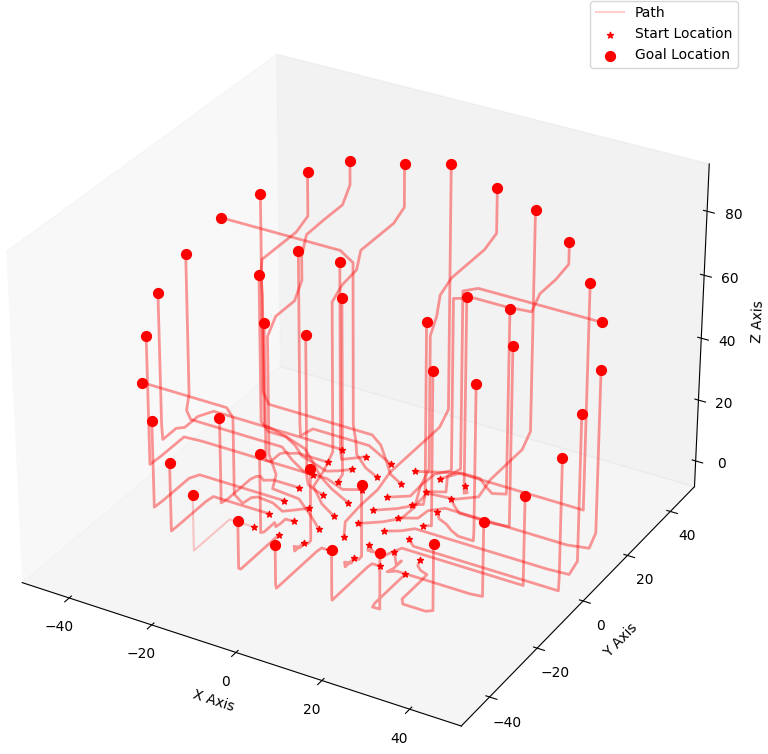}
    \caption{`Smiley Face' formation in a 45-degree slant using 41 UAVs. Starting and goal positions are represented using stars and dots, respectively.}
    \label{fig:smiley}
\end{figure}

\begin{figure}[!ht]
    \centering
    \includegraphics[width=0.7\linewidth]{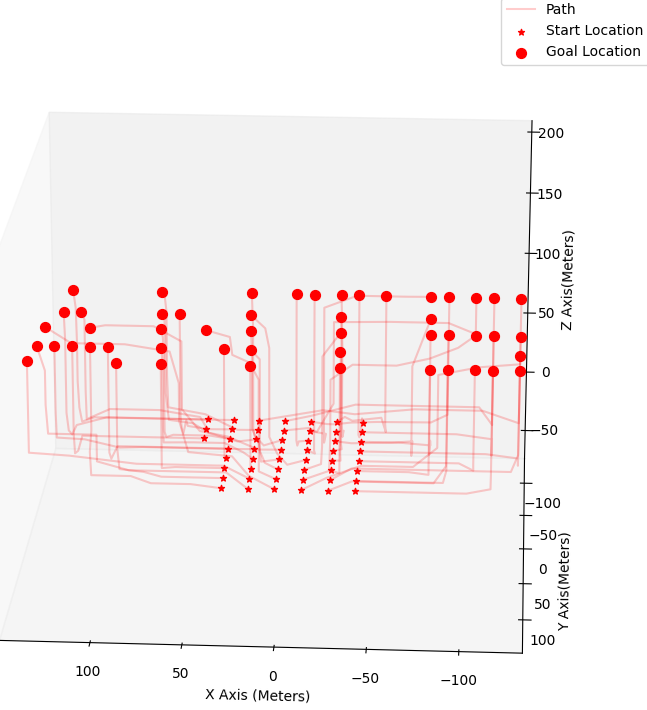}
    \caption{`ANTS' formation using 51 UAVs}
    \label{fig:ants}
\end{figure}

\begin{figure}[!ht]
    \centering
    \includegraphics[width=0.75\linewidth]{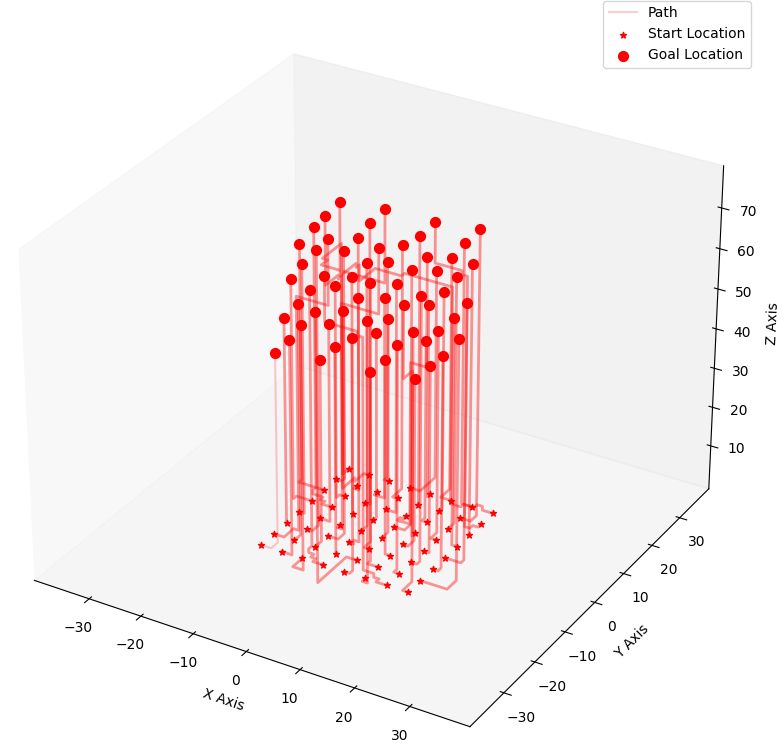}
    \caption{‘Cube’ formation using 64 UAVs}
    \label{fig:cube}
\end{figure}

\section{Conclusion and Future works}
In this paper, we have implemented a method of utilizing graph-based flow networks to devise a method for MFF. It has three main parts: first, it generates a flow network graph from physical GPS coordinates; second, it uses the Bellman-Ford algorithm to find the path with the shortest distance; and finally, it employs the Ford-Fulkerson Method to find collision-free paths. We can ensure that it generates paths in which UAVs can safely fly in close proximity to one another. It was validated by several software-in-the-loop (SITL) simulations of up to 64 UAVs, as well as by conducting multiple outdoor flight tests in predefined formations of 3 quadcopters. Fast computation times for finding the paths for MFF demonstrated that our method is efficient and reliable.

Future improvements to this work could be in terms of improvement in localization techniques, as GPS-based localization can be expensive and imprecise at times. We may also incorporate a trajectory planner in future to allow for a more seamless and coordinated movement of UAVs.

Furthermore, the addition of velocity control will enable us to perform direct comparisons with current methods of MFF. This is due to the fact that a time-based parameter is typically used in path or trajectory planning literature as a performance indicator.

\begin{figure}[!htb]
    \centering
    \includegraphics[width=\linewidth]{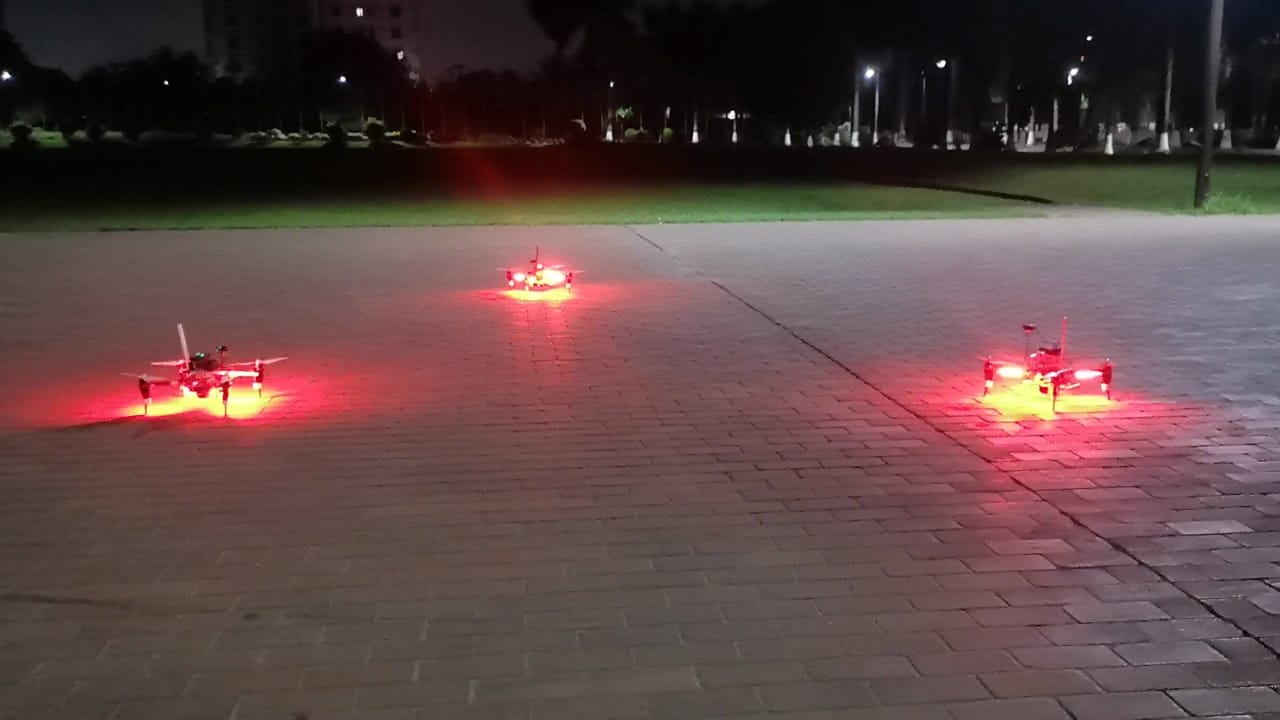}
    \caption{The quadcopters (UAVs) in their starting positions.}
    \label{fig:real1}
\end{figure}

\begin{figure}[!htb]
    \centering
    \includegraphics[width=\linewidth]{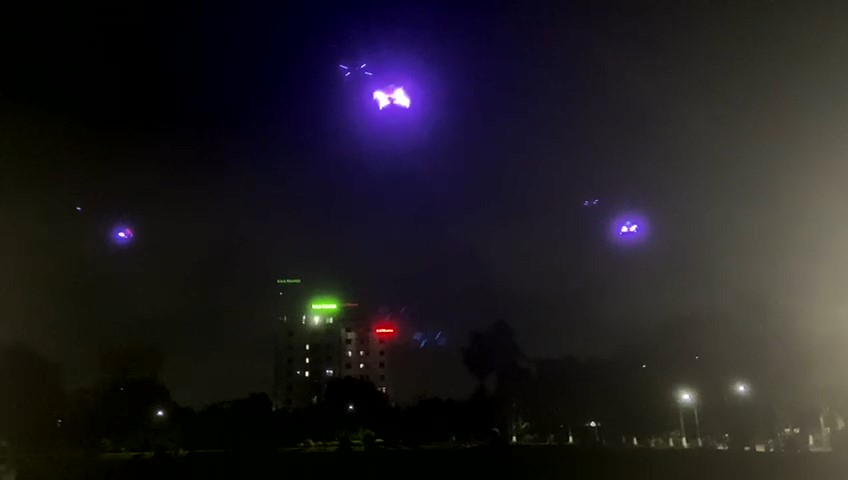}
    \caption{Quadcopters in operation, following the paths generated by the MFF Solver.}
    \label{fig:real2}
\end{figure}

\printbibliography
\end{document}